\documentclass[english]{article}
\usepackage{spconf,amsmath,balance}
\usepackage[T1]{fontenc}
\usepackage[latin9]{inputenc}
\usepackage{amsthm}
\usepackage{amsmath}
\usepackage{amssymb}
\usepackage{graphicx}
\usepackage{balance}
\usepackage{epstopdf}
\usepackage{algpseudocode}
\usepackage{algorithm}

\makeatletter

\theoremstyle{plain}

\theoremstyle{remark}

\makeatother

\usepackage{babel}
\providecommand{\remarkname}{Remark}
\providecommand{\theoremname}{Theorem}
\graphicspath{{Figures/}}

\begin{document}

\title{Frequency estimation in three-phase power systems\\ \vspace{0.2cm} with harmonic contamination: \\ \vspace{0.2cm} A multistage quaternion Kalman filtering approach}

\name{Sayed Pouria Talebi and Danilo P. Mandic}
\address{Imperial College London, Department of Electrical and Electronic Engineering
\\E-mails: \{s.talebi12, d.mandic\}@imperial.ac.uk}

\author{Sayed Pouria Talebi and Danilo P. Mandic}

\maketitle
\begin{abstract}
Motivated by the need for accurate frequency information, a novel algorithm for estimating the fundamental frequency and its rate of change in three-phase power systems is developed. This is achieved through two stages of Kalman filtering. In the first stage a quaternion extended Kalman filter, which provides a unified framework for joint modeling of voltage measurements from all the phases, is used to estimate the instantaneous phase increment of the three-phase voltages. The phase increment estimates are then used as observations of the extended Kalman filter in the second stage that  accounts for the dynamic behavior of the system frequency and simultaneously estimates the fundamental frequency and its rate of change. The framework is then extended to account for the presence of harmonics. Finally, the concept is validated through simulation on both synthetic and real-world data.  
\end{abstract}

\section{Introduction}

The power grid is designed to operate optimally at its nominal frequency; furthermore, deviations from the nominal system frequency adversely affects the components of the power grid~\cite{ModelingEffects}, such as compensators and loads, resulting in harmful operating conditions that can propagate throughout the network. Thus, making frequency stability as one of the most important factors in the assessment of power quality~\cite{PowerQuality}. In addition, future envisioned smart grid technologies will incorporate distributed power generation based on renewable energy sources, where the wide-area grid can be divided into a number of smaller self-contained sections named micro-grids, with some micro-grids disconnecting from the wide-area grid for prolonged lengths of time, referred to as islanding~\cite{Mag}.  Therefore, monitoring the frequency information of the grid in real-time is an essential part of power distribution network control and management applications.

The importance of frequency estimation in power grids has motivated the development of a variety of algorithms dedicated to this cause~\cite{RTFE}-\cite{Me}, where approaches based on Kalman filtering have been shown to be advantageous due to their underlying state space assumption~\cite{CEKF}-\cite{Me}. However, these algorithms act as frequency measurement techniques and do not reveal any information about the dynamics of the system frequency which holds essential information about the power grid. For example, a rising system frequency indicates that power generation has surpassed power consumption and a falling system frequency is indicative that power consumption has exceeded power generation~\cite{Mag},\cite{FRC}. 

Although the voltage measurements from the three phases can be mapped onto the complex domain using the Clarke transform, which allows for the use of well established complex-valued signal processing techniques, complex numbers lack the dimensionality necessary to model three-phase systems leading to partial loss of information specially under unbalanced operating conditions which negatively effects the performance complex-valued frequency estimators~\cite{Me}. Quaternions provide a natural presentation for three and four-dimensional signals and have gained popularity in an increasing number of engineering applications~\cite{Me}-\cite{FAT}. In addition, the recent introduction of the $\mathbb{HR}$-calculus~\cite{HR} and augmented quaternion statistics~\cite{Took}-\cite{Via} have led a resurgence in quaternion-valued signal processing and have inspired a number of quaternion-valued signal processing algorithms including a class of quaternion Kalman filters~\cite{QKF}.

In this work, a novel two stage Kalman filtering algorithm is developed for adaptive estimation of frequency and its rate of change in three-phase power systems. The first stage is consisted of a strictly linear quaternion extended Kalman filter (QEKF) that allows for the incorporation of the voltage measurements from all phases and estimates the instantaneous phase increment of the three-phase voltages. The instantaneous phase increment estimates from the first stage are then used as observations of a real-valued extended Kalman filter (EKF) in the second stage that estimates the system frequency and its rate of change. Moreover, a framework for accounting for the presence of harmonics in the power system is presented. The concept is verified through simulations on both synthetic data mirroring practical power grid scenarios and real-world data recordings.

\section{Quaternions}

Quaternions are a non-commutative division algebra denoted by $\mathbb{H}$. A quaternion variable $q \in \mathbb{H}$ consists of a real part $\Re(q)$ and a three-dimensional imaginary part or pure quaternion, $\Im(q)$, which comprises three imaginary components $\Im_{i}(q)$, $\Im_{j}(q)$, and $\Im_{k}(q)$; therefore, $q$ can be expressed as $q =\Re(q) + \Im(q) =\Re(q) + \Im_{i}(q) + \Im_{j}(q) + \Im_{k}(g) = q_{r} + iq_{i} + jq_{j} + kq_{k}$, where $q_{r},q_{i},q_{j},q_{k} \in \mathbb{R}$, while $i$, $j$, and $k$ are imaginary units obeying the product rules $ij=k$, $jk=i$, $ki=j$, and $i^{2}=j^{2}=k^{2} = ijk = -1$.

Alternatively a quaternion $q \in \mathbb{H}$ can be expressed by the polar presentation given by \cite{QFFT}
\[
q = |q|e^{\xi \theta}=|q|\big(\text{cos}(\theta) + \xi \text{sin}(\theta) \big)
\]
where
\[
\xi = \frac{\Im(q)}{|\Im(q)|} \hspace{0.5em}\text{and}\hspace{0.5em} \theta = \text{atan}\left(\frac{|\Im(q)|}{\Re(q)}\right).
\]
Moreover, it is straightforward to prove that the $\text{sin}(\cdot)$ and $\text{cos}(\cdot)$ functions can be expressed as
\[
\text{sin}(\theta)=\frac{1}{2\xi}\left(e^{\xi \theta} - e^{- \xi \theta}\right), \hspace{0.5em}\text{cos}(\theta)=\frac{1}{2}\left(e^{\xi \theta} + e^{- \xi \theta}\right)
\label{eq:quaternion sin and cos}
\]
where $\xi^{2} = -1$.

The involution of $q\in \mathbb{H}$ around $\eta \in \mathbb{H}$ is defined as $q^{\eta} \triangleq \eta q \eta^{-1}$~\cite{QIAI} and can be used to express the real-valued components of a quaternion number, $q \in \mathbb{H}$, as \cite{FAT}-\cite{Took},\cite{QKF}
\begin{equation}
\begin{aligned}
q_{r} =& \frac{1}{4}\left(q + q^{i} + q^{j} + q^{k} \right) & q_{i} =&\frac{1}{4i}\left(q + q^{i} - q^{j} - q^{k}\right)
\end{aligned}
\label{eq:real-valued components}
\end{equation}
\[
\begin{aligned}
q_{j} =& \frac{1}{4j}\left(q - q^{i}+ q^{j} - q^{k} \right) & q_{k} =&\frac{1}{4k}\left(q - q^{i} - q^{j} + q^{k}\right).
\end{aligned}
\]
The quaternion conjugate is a special case of quaternion involutions and is defined as
\begin{equation}
q^{*}=\Re(q)-\Im(q)=\frac{1}{2}\left(q^{i} + q^{j} + q^{k} - q \right)
\label{eq:quaternion conjugate}
\end{equation}
while the norm of $q \in \mathbb{H}$ is given by 
\[
\left|q\right|=\sqrt{qq^{*}}=\sqrt{ q^{2}_{r} + q^{2}_{i} + q^{2}_{j} + q^{2}_{k} }.
\]
The expressions in (\ref{eq:real-valued components}) establish a relation between the augmented quaternion variable, $[\mathbf{q},\mathbf{q}^{i},\mathbf{q}^{j},\mathbf{q}^{k}]^{T} \in \mathbb{H}^{4}$, and the real-valued vector $[\mathbf{q}_{r},\mathbf{q}_{i},\mathbf{q}_{j},\mathbf{q}_{k}]^{T} \in \mathbb{R}^{4}$. This duality between the real and quaternion domains forms the basis of the quaternion augmented statistics~\cite{Took}-\cite{Via} and the $\mathbb{HR}$-calculus~\cite{HR}. The augmented quaternion statistics in conjunction with the $\mathbb{HR}$-calculus have led to the development of a class of quaternion Kalman filters including the strictly linear QEKF~\cite{QKF} that operates akin to its complex-valued counterpart, with the difference that the Jacobian of the state evolution function is calculated by the $\mathbb{HR}$-calculus. For instance, $\frac{\partial q^{*}}{ \partial q}=-0.5$, which is a consequence of (\ref{eq:quaternion conjugate}) and is in contrast with the results in the complex domain. 

\section{Quaternion Frequency Estimator}

The instantaneous voltages of each phase in a three-phase power system are given by
\begin{equation}
\begin{aligned}
v_{a,n}=&V_{a,n}\text{sin}\left(2\pi f_{n} \Delta T n + \phi_{a,n}\right)
\\
v_{b,n}=&V_{b,n}\text{sin}\Big(2\pi f_{n} \Delta T n +  \phi_{b,n} +  \frac{2 \pi}{3}\Big)
\\
v_{c,n}=&V_{c,n}\text{sin}\Big(2\pi f_{n} \Delta T n +  \phi_{c,n} +  \frac{4 \pi}{3}\Big)
\end{aligned}
\label{eq:phase voltages}
\end{equation}
where $V_{a,n}$, $V_{b,n}$, and $V_{c,n}$ are the instantaneous amplitudes, $\phi_{a,n}$, $\phi_{b,n}$, and $\phi_{c,n}$ represent the instantaneous phases, and $\Delta T = 1/f_{s}$ is the sampling interval with $f_{s}$ denoting the sampling frequency, while $f_{n}$ denotes the system frequency at time instant $n$. Considering that the dynamic changes in system frequency can be modeled as a time varying function given by $f_{n}=g(n\Delta T)$; then, taking the Taylor series expansion yields
\begin{equation}
\begin{aligned}
f_{n+1}=&\underbrace{g (n\Delta T)}_{f_{n}}+\Delta T \underbrace{\left.\frac{\partial g}{\partial t}\right|_{t=n\Delta T}}_{r_{n}}
\\
&+\underbrace{\sum^{\infty}_{l=2}\Delta T^{l} \left.\frac{\partial^{l}g}{\partial t^{l}}\right|_{t= n \Delta T}}_{\text{h.o.t}}
\end{aligned}
\label{eq:Freq-Taylor}
\end{equation}
where $r_{n}$ is the rate of change of the system frequency, while the higher order terms are denoted by $\text{h.o.t}$ and are assumed to be sufficiently small for the system frequency to be approximated by the first order regression\footnote{The framework developed in this work can be expanded to account for additional terms of the Taylor series expansion in (\ref{eq:Freq-Taylor}) if further information about the dynamics of the system frequency is required.}
\begin{equation}
f_{n+1}=f_{n}+r_{n}\Delta T.
\label{eq:Freq-Rec}
\end{equation}
Taking into account the linear regressions in (\ref{eq:Freq-Rec}) a state space model for the system frequency that can be implemented using the real-valued EKF is proposed in Algorithm~\ref{Al:F-SS}, where $\boldsymbol{\nu}_{n}$ and $\omega_{n}$ denote the state evolution and observation noise.
\begin{algorithm}
\caption{State space model of system frequency (F-SS)} \vspace{0.1cm}
State evolution equation: $\begin{bmatrix} r_{n+1}\\ f_{n+1}\end{bmatrix}=\begin{bmatrix}r_{n}\\ f_{n} + r_{n}\Delta T \end{bmatrix}+\boldsymbol{\nu}_{n}$ \vspace{0.5cm}
\\
Observation equation: $f_{n}+r_{n}\Delta T + \omega_{n}$ \vspace{0.5em}
\label{Al:F-SS}
\end{algorithm}

The three phase voltages in (\ref{eq:phase voltages}) are now combined together to generate the pure quaternion signal
\begin{equation}
q_{n}=iv_{a,n}+jv_{b,n}+kv_{c,n}.
\label{eq:quaternion signal}
\end{equation}
Since all three phases have the same frequency, analytical geometry dictates that $q_{n}$ will trace an ellipse in a subspace (one plane) of the three-dimensional imaginary subspace of~$\mathbb{H}$~\cite{AG}. In order to simplify our analysis we consider the the set of imaginary bases $\{\zeta, \zeta',\zeta''\}$ such that
\begin{equation}
\begin{array}{c}
\zeta \zeta' = \zeta'',\hspace{0.5em} \zeta' \zeta'' =\zeta,\hspace{0,5em} \zeta'' \zeta = \zeta'
\end{array}
\label{eq:new imaginary units}
\end{equation}
and the imaginary units $\{\zeta,\zeta'\}$ are designed to reside in the same plane as $q_{n}$, resulting in $\zeta''$ being normal to this plane. In our previous work we show that an arbitrary ellipse in the $\zeta-\zeta'$ plane can then be divided into two counter rotating circular components $q^{+}_{n}$ and $q^{-}_{n}$, which can be expressed by the corresponding fist order quaternion linear regressions~\cite{Me}
\begin{equation}
q^{+}_{n+1}= e^{\zeta'' \Delta \theta_{n}}q^{+}_{n} \hspace{0.5em} \text{and} \hspace{0.5em} q^{-}_{n+1}= e^{-\zeta'' \Delta \theta_{n}}q^{-}_{n}
\label{eq:quaternion regressive}
\end{equation}
where $\Delta \theta_{n} = 2 \pi f_{n} \Delta T + 2 \pi r_{n} \Delta T^{2}$ is the phase increment of $q^{+}_{n}$ and $q^{-}_{n}$ between time instances $n$ and $n+1$.

Taking into account the linear regressions in (\ref{eq:quaternion regressive}), a state space model for $q_{n}$ is proposed in Algorithm~\ref{Al:Q-SS}, where $\varphi_{n} = e^{\zeta'' \Delta \theta}$, $\boldsymbol{\upsilon}_{n}$ and $\epsilon_{n}$ are the state evolution and observation noise. Note that Algorithm~\ref{Al:Q-SS} can be implemented using the strictly linear QEKF presented in~\cite{QKF}.
\begin{algorithm}
\caption{Quaternion-valued state space model (Q-SS)}\vspace{0.1cm}
State evolution equation: $\begin{bmatrix}\varphi_{n+1}\\ q^{+}_{n+1}\\ q^{-}_{n+1}\end{bmatrix}=\begin{bmatrix}\varphi_{n}\\ \varphi_{n} q^{+}_{n}\\ \varphi^{*}_{n} q^{-}_{n}\end{bmatrix}+\boldsymbol{\upsilon}_{n}$ \vspace{0.5em}
\\
Observation equation: $q_{n}=\begin{bmatrix}0&1&1\end{bmatrix}\begin{bmatrix}\varphi_{n}\\ q^{+}_{n}\\ q^{-}_{n}\end{bmatrix}+ \epsilon_{n}$ \vspace{0.5em}
\\  \vspace{5pt}
Estimate of phase increment: $\widehat{ \Delta \theta_{n} }= \Im \left( \text{ln}\left(\varphi_{n} \right) \right)$\\ 
\vspace{-8pt}
\label{Al:Q-SS}
\end{algorithm}

\begin{figure}[!h]
\centering
\includegraphics[width=1\linewidth,  trim = 0cm 0.5cm 0cm 0cm]{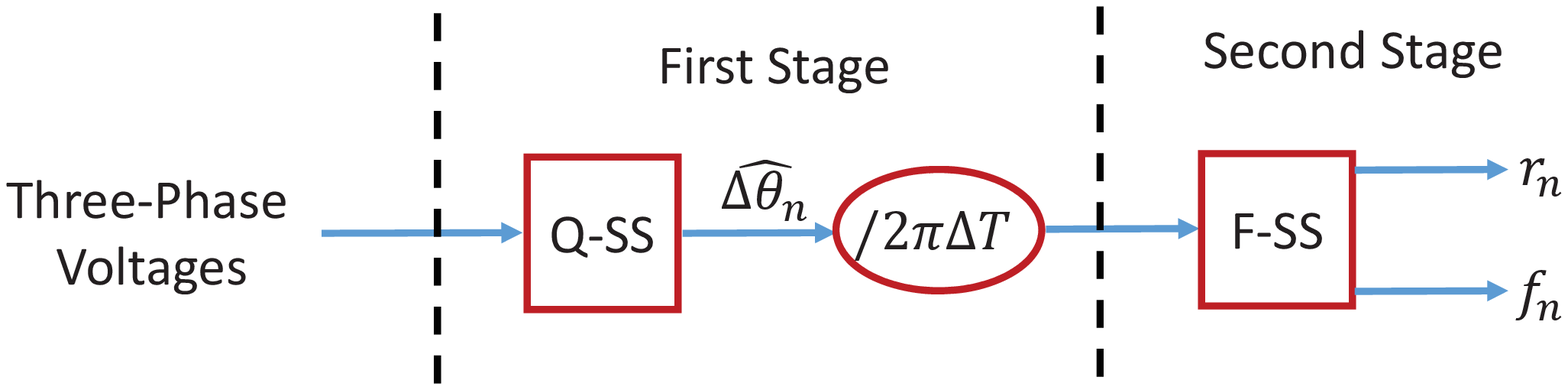}
\caption{Schematic of the developed two stage estimator of frequency and its rate of change.}
\label{fig:schematic1}
\end{figure}

Since the Kalman filter is an unbiased estimator, it is straightforward to show $\widehat{\Delta \theta_{n}}$ is an unbiased estimate of the phase increment; furthermore, assuming that $\boldsymbol{\upsilon}_{n}$ and $\omega_{n}$ are Gaussian variables, makes it possible to model the estimate of the phase increment as $\widehat{\Delta \theta_{n}}=\Delta \theta_{n} + \mu_{n}$, where $\mu_{n}$ is real-valued white Gaussian noise. Therefore, the relation between the phase increment, estimated in Algorithm~\ref{Al:Q-SS} and the observation equation in Algorithm~\ref{Al:F-SS} given by
\[
f_{n}+r_{n}\Delta T + \omega_{n} = \frac{\widehat{\Delta \theta}_{n}}{2\pi \Delta T}=\frac{\Delta \theta_{n}+\mu_{n}}{2\pi\Delta T}
\]
allows the system frequency and its rate of change to be estimated simultaneously through the use of Algorithm~\ref{Al:F-SS} in series with Algorithm~\ref{Al:Q-SS} in the fashion shown in Figure~\ref{fig:schematic1}.

\begin{figure}[!h]
\centering
\includegraphics[width=1\linewidth,  trim = 0cm 0.2cm 0cm 0cm]{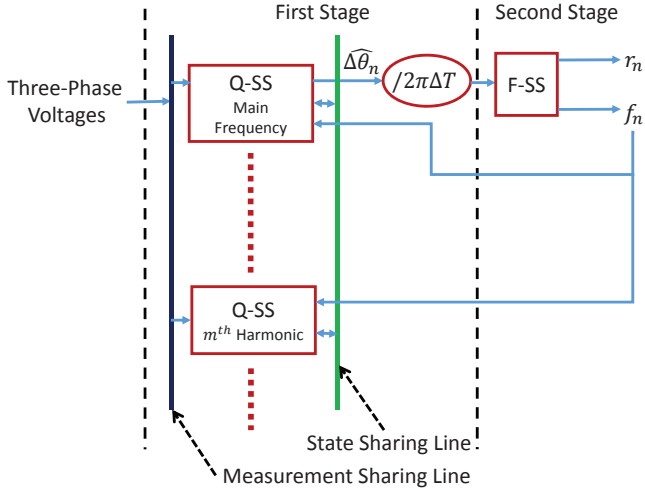}
\caption{Schematic of the developed two stage estimator of frequency and its rate of change with Multi-QEKF in the first stage accounting for the presence of harmonics.}
\label{fig:schematic2}
\end{figure}

Harmonic contamination is a major cause of error in power system frequency estimators and therefore it is crucial to present a framework for accounting for the presence of the main harmonic components. In this setting, the estimate of the system frequency is used as feedback in order to estimate the signal component at $m$-times the main frequency. This is achieved through the use of multiple QEKF connected in parallel as shown in Figure~\ref{fig:schematic2}, where each QEKF is implementing the state space model in Algorithm~\ref{Al:Q-SS}. The process is summarized in Algorithm~\ref{Al:QFE}, where $\hat{\mathbf{x}}_{m,n|n-1}$, $\hat{\mathbf{x}}_{m,n|n}$, $\mathbf{A}_{m,n}$, and $\mathbf{C}_{\boldsymbol{\upsilon}_{m,n}}$ are the \textit{a priori} state estimate, \textit{a posteriori} state estimate, the Jacobian of the state evolution equation, and the covariance matrix of the state evolution noise at time instant $n$ for the QEKF estimating the $m^{\text{th}}$ harmonic component, while $\mathbf{C}_{\boldsymbol{\upsilon}_{m,n}}$ is covariance matrix of the observation noise, with $\mathbf{h}=[0\text{ }1\text{ }1]$ denoting the observation function.
\begin{algorithm}
\caption{Quaternion Frequency Estimator (QFE)}\vspace{0.1cm}
Initiate: $\hat{\mathbf{x}}_{m,0|0}$ and $\mathbf{P}_{m,0|0}$ for $m=1,3,5,\ldots$.
\\ 
Compute the model output: \vspace{-0.1cm}
\[
\begin{aligned}
\hat{\mathbf{x}}_{m,n|n-1}&=\mathbf{A}_{m,n}\hat{\mathbf{x}}_{m,n-1|n-1}
\\
\mathbf{P}_{m,n|n-1}&= \mathbf{A}_{m,n}\mathbf{P}_{m,n-1|n-1}\mathbf{A}^{H}_{m,n}+\mathbf{C}_{\boldsymbol{\upsilon}_{m,n}}
\end{aligned}
\] \vspace{-0.1cm}
\\ 
Share the \textit{a priori} state estimate $\hat{\mathbf{x}}_{m,n|n-1}$.
\\
Compute the measurement output: \vspace{-0.1cm}
\[
\begin{aligned}
\mathbf{G}_{m,n}&=\mathbf{P}_{m,n|n-1}\mathbf{h}^{H}\big(\mathbf{h}\mathbf{P}_{m,n-1|n-1}\mathbf{h}^{H}+\mathbf{C}_{\omega_{n}} \big)^{-1}
\\
\hat{\mathbf{x}}_{m,n|n}&=\hat{\mathbf{x}}_{l,n|n-1}+\mathbf{G}_{m,n}\big(q_{n}-\mathbf{h}\text{\hspace{-0.1cm}}\sum_{l=1,3,\ldots}\text{\hspace{-0.2cm}}\hat{\mathbf{x}}_{l,n|n-1} \big)
\\
\mathbf{P}_{m,n|n}&=\big(\mathbf{I}-\mathbf{G}_{m,n}\mathbf{h}\big)\mathbf{P}_{m,n|n-1}
\end{aligned}
\]\vspace{-0.1cm}
\\
Save the phase incrementing element, $\varphi_{m,n|n-1}=e^{\xi''_{m,n} \widehat{\Delta \theta}_{n}}$.
\\
Estimate the system frequency, $f_{n}$, in the second stage.
\\
Update the phase incrementing element: \vspace{-0.1cm} 
\[
\varphi_{m,n|n}=e^{\xi''_{m,n} m f_{n} \Delta T}
\]
\vspace{-0.5cm}
\label{Al:QFE}
\end{algorithm}

\vspace{-0.2cm}

\section{Simulations}

The performance of the newly developed algorithm was validated in different experiments, where the sampling frequency was $f_{s} = 1\text{kHz}$ and the voltage measurements were considered to be corrupted by white Gaussian noise with signal to noise ratio of $30$dB.

\begin{figure}[!h]
\centering
\includegraphics[width=1\linewidth, trim = 0cm 1cm 0cm 0cm]{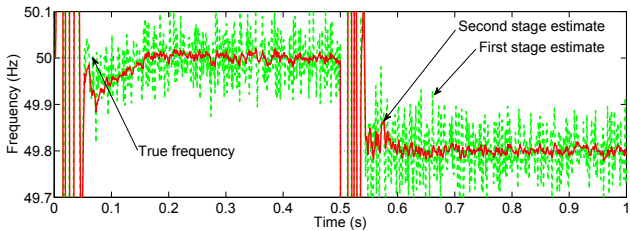}
\caption{Frequency estimation for a three-phase system experiencing a voltage sag and a $10$\% third harmonic from $0.5$ seconds after simulation started.}
\label{fig:jump}
\end{figure}

In the first experiment, the three-phase system was considered to be operating under balanced conditions, at its nominal frequency of $50\text{Hz}$, and free of harmonics; then suffered a voltage sag characterized by an $80$\% fall in the amplitude of $v_{a}$ and $20$ degree phase shifts in $v_{b}$ and $v_{c}$; furthermore, the voltage sag resulted in the introduction of a balanced $10$\% third harmonic component and a $0.2$Hz fall in system frequency. The estimates of the system frequency obtained at the first and second stages of the developed algorithm are shown in Figure~\ref{fig:jump}. Notice that the developed algorithm accurately estimated the system frequency under both balanced and unbalanced conditions regardless of the presence of the third harmonic.   

\begin{figure}[!h]
\centering
\includegraphics[width=1\linewidth, trim = 0cm 0cm 0cm 0cm]{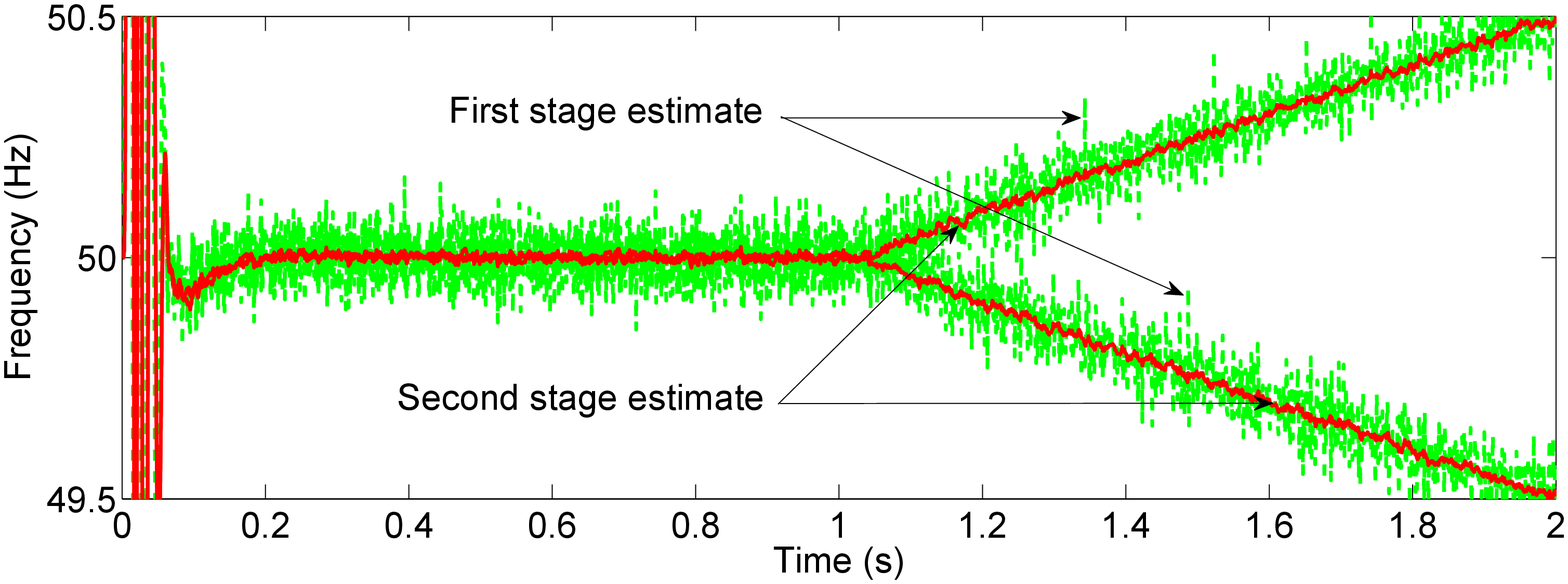}
\\
\includegraphics[width=1\linewidth, trim = 0cm 1.2cm 0cm 0cm]{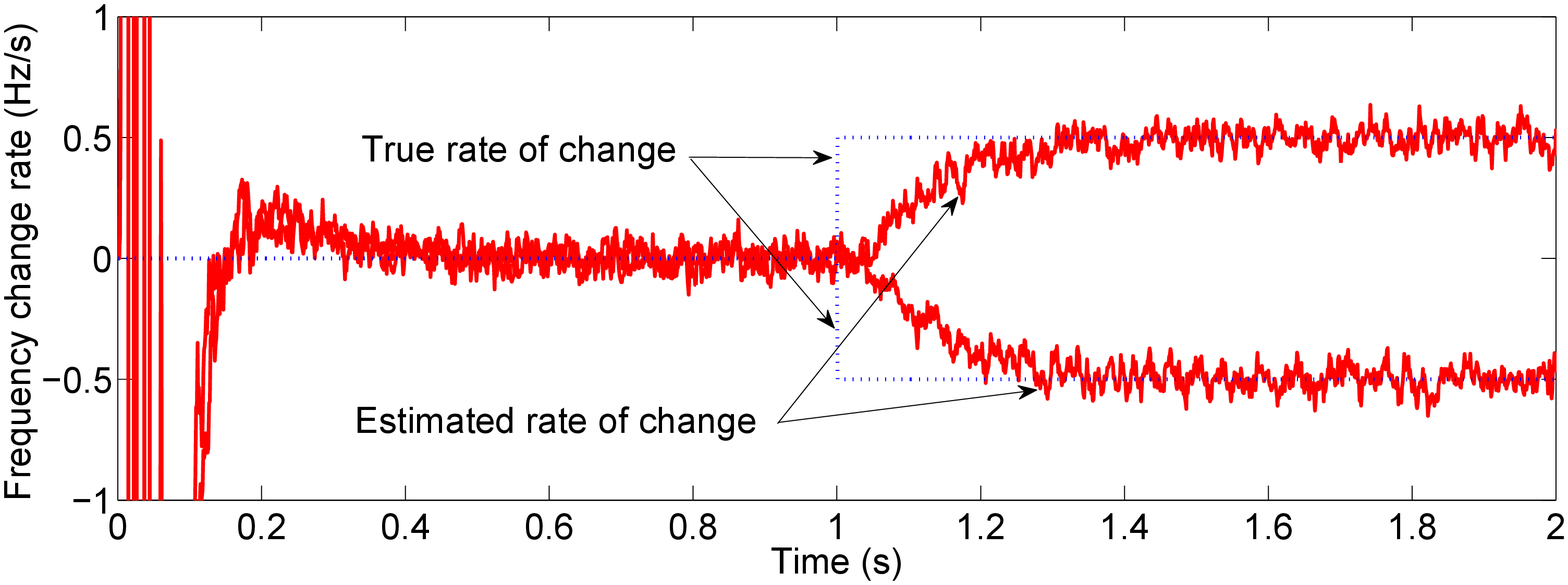}
\caption{Frequency estimation for an unbalanced three-phase system experiencing frequency rise or fall while suffering from a $10$\% third harmonic. The system frequency is shown in the top figure and the frequency rate of change is shown in the bottom figure.}
\label{fig:ramp}
\end{figure}

In the second experiment, a three-phase system operating under the same unbalanced condition and harmonic contamination as the previous experiment, experienced a rising (\textit{cf}. falling) frequency at the rate of $0.5\text{Hz/s}$ due to mismatch between power generation and consumption $0.5$ seconds after simulation started. The estimates of the system frequency and its rate of change are shown in Figure~\ref{fig:ramp}. Observe that the algorithm accurately estimated the system frequency and its rate of change. 

Finally, the performance of the developed quaternion frequency estimator, QFE, is compared to the recently introduced complex-valued frequency estimator in~\cite{D.H}, CFE,  using real-world data. Notice that although both algorithms obtain reliable estimates of system frequency, the QFE achieves a smaller steady state variance.   
\begin{figure}[!h]
\centering
\includegraphics[width=1\linewidth,  trim = 0cm 1.2cm 0cm 0.4cm]{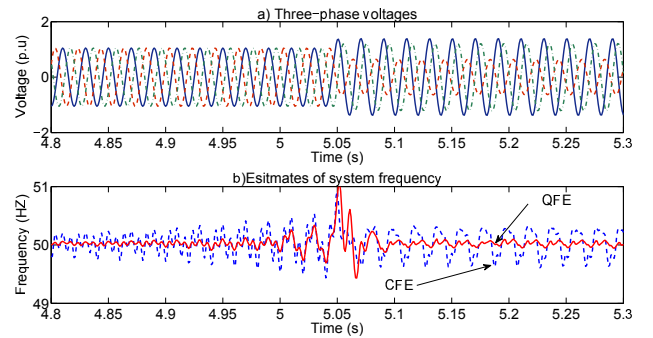}
\caption{Frequency estimation during a real-world voltage sag: a) system voltages, b) estimates of the system frequency obtained by the QFE and CFE algorithms.}
\end{figure}

\vspace{-0.5cm}

\section{Conclusion}

A novel frequency estimation algorithm for three-phase power systems has been developed based on a two stage Kalman filtering approach. The proposed algorithm has been shown to fully characterize both balanced and unbalanced three-phase power systems and account for the presence of harmonics. The performance of the proposed algorithm has been assessed in a number of scenarios using synthetic data and shown to outperform conventional complex frequency estimators using real-world data recoding.

\balance

\end{document}